\pdfoutput=1

\documentclass[11pt]{article}

\usepackage{EACL2023}

\usepackage{annotates}
\usepackage{times}
\usepackage{latexsym}

\usepackage[T1]{fontenc}

\usepackage[utf8]{inputenc}

\usepackage[english]{babel}
\usepackage[english=british]{csquotes}

\usepackage{microtype}

\usepackage{graphicx}
\usepackage{booktabs}
\usepackage{float}
\usepackage{amsmath}
\usepackage{paralist}

\newcommand{\MP}{MPNET}
\newcommand{\DR}{DistilRoberta}
\newcommand{\BERT}{BERT}

\title{Representation biases in sentence transformers}

\author{Dmitry Nikolaev \qquad Sebastian Pad{\'o} \\
  IMS, University of Stuttgart \\
  \texttt{dnikolaev@fastmail.com} \qquad \texttt{pado@ims.uni-stuttgart.de}
  }

\begin{document}
\maketitle

\begin{abstract}
  Variants of the BERT architecture specialised for producing
  full-sentence representations often achieve better performance on
  downstream tasks than sentence embeddings extracted from vanilla
  BERT. However, there is still little understanding of what
  properties of inputs determine the properties of such
  representations. In this study, we construct several sets of
  sentences with pre-defined lexical and syntactic structures and show
  that SOTA sentence transformers have a strong
  nominal-participant-set bias: cosine similarities between pairs of
  sentences are more strongly determined by the overlap in the set of
  their noun participants than by having the same predicates, lengthy
  nominal modifiers, or adjuncts. At the same time, the precise
  syntactic-thematic functions of the participants are largely
  irrelevant.
\end{abstract}
\section{Introduction}

Transformer-based encoder-only models derived from the BERT
architecture and pre-trained using similar objective and training
regimens \cite{devlin-etal-2019-bert,liu2019roberta} have become the
standard tool for downstream tasks at the level of individual tokens
and token sequences \citep{tenney-etal-2019-bert,wang2021ner}.
Whole-sentence representations can also be easily extracted from the
outputs of these models by either using the embedding of the special
[CLS] token, in cases where the model was trained on the
next-sentence-prediction task, or averaging or max-pooling the
embeddings of all tokens produced by the model
\citep{zhelezniak2019max}. While both approaches are widely used in
practice, it has been argued that these representations are not well suited
for sentence-level downstream tasks. Several modifications to the
architecture and training regime were proposed, which  are known collectively as sentence transformers
\citep[STs;][]{reimers-gurevych-2019-sentence}.

STs have achieved state-of-the-art performance on downstream tasks
such as semantic search and question answering
\citep{santander2022semantic,ha2021utilizing}. Their analysis,
however, has received considerably less attention than the analysis of
the vanilla BERT model and its variants
\citep{rogers-etal-2020-primer,conia-navigli-2022-probing}. In fact, these models are often considered to be uninterpretable
\citep{minaee2021deep}.

A common feature of STs is that they are fine-tuned
to produce similar vector-space representations for semantically
similar sentences.
%
%
This objective induces a complex loss landscape shaped by the available training data.
The original Sentence-BERT model
\citep{reimers-gurevych-2019-sentence} was trained on
natural language inference data, and sentences were considered to be
semantically similar if their NLI label was that of
entailment. SOTA models were trained on a much larger
web-crawled corpus including more than 1 billion sentence pairs mined
from sources such as Reddit conversations, duplicate question pairs
from WikiAnswers, etc.\footnote{See the list at
\url{https://huggingface.co/sentence-transformers/all-mpnet-base-v2}}
The richness and variability of this dataset begs the question of
what notion of semantic similarity is implicitly learned by the models
trained on it.

In this study, we begin addressing this question through analysis of
natural-looking synthetic sentences with controlled syntactic and lexical
content. We concentrate on three questions.


First, we test if STs have part-of-speech biases. We show that,
all other things being equal, information provided by nouns plays more important
role than the information provided by verbs, both in simple sentences and in sentences
with coordinated verbal phrases. 

Second, we compare the relative importance of the overlap in the sets of participants 
in two sentences with that of how many participants have identical syntactic
functions. We show that raw lexical overlap is relatively more important than having
the same nouns in the same syntactic slots.

Third, we check how strongly sentence representations are affected by other
sentential elements, such as adverbials and nominal modifiers of different types
and lengths. We show that, unlike BERT with token averaging, STs
seem to largely disregard these components in favor of nominal participants.

The paper is structured as follows: \S~\ref{sec:methods} presents the
methodology that we follow in our analyses and the models we employ;
\S~\ref{sec:case-studies} presents the case studies and their
results; \S~\ref{sec:discussion} provides an overall discussion;
\S~\ref{sec:related-work} surveys related work;
\S~\ref{sec:conclusion} concludes the paper.

\section{Methods and Experimental Setup}\label{sec:methods}

We experiment with representations produced by three models.
Two are SOTA STs:
\texttt{all- mpnet-base-v2} (\MP) is an instance of \texttt{mpnet-base}
\citep{song2020mpnet} fine-tuned on the 1B sentence-pair corpus using
the training architecture from \citet{reimers-gurevych-2019-sentence};
\texttt{all-distilroberta-v1} (\DR) is a distilled instance of
\texttt{roberta-base} \citep{Sanh2019DistilBERTAD} fine-tuned in the
same way. The third model is the vanilla pre-trained
\texttt{bert-large-uncased} (\BERT), as a point of comparison for the first two.

All models were downloaded from HuggingFace.
Standard APIs from the Sentence Transformers 
library\footnote{\url{https://www.sbert.net/index.html},
\citet{reimers-gurevych-2019-sentence}.} were used to compute embeddings using
\MP\ and \DR; for the vanilla BERT model, we averaged the embeddings of all sentence tokens,
including [CLS] and [SEP].\footnote{We
experimented with omitting the special tokens, but this led to sentence representations
dominated by punctuation signs and other undesired effects. In line with previous work
\citep{ma2019universal}, we also found that using [CLS] embeddings leads to
bad results due to their high redundancy, and we do not discuss them.}

We structure the presentation as a series of case studies. For each
case study, we construct a set of sentences controlled for
lexical content and syntactic structure. Sentences are created in such
a way as to be grammatically correct, look naturalistic, and as far as
possible not bias the analysis.\footnote{Sentence-generating and model-fitting scripts
can be found in the Supplementary Materials.} They are arguably less complex and variable than 
examples sampled from real-word corpora; however, we believe that an analysis based
on simple sentences is a reasonable first step towards a better understanding of model
representations, as previous work has shown for sentiment analysis \citep{kiritchenko-mohammad-2018-examining} and syntactic analysis \citep{marvin-linzen-2018-targeted}.

For each case study, we compute embeddings for all sentences, together
with cosine similarities between embeddings of sentence pairs. We
 analyze the similarities by means of regression modelling. More
precisely, we regress cosine similarities,
z-scored to improve comparability between encoders, on the properties
of sentence pairs, such as lexical overlap, presence of identical
participants in identical syntactic positions, or POS tags of
participants.
We  inspect the
coefficients of the resulting regression fits to assess the relative
importance of these properties. Since (almost) all  properties are coded as binary variables, their magnitudes are directly comparable in terms of importance.

For terminological clarity, we will use the term \textit{models} to refer to the
regression models we use to analyse the impact of sentence properties on representational
similarity. We call the transformers computing these embeddings \textit{encoders}.

Where the features of sentence pairs can be straightforwardly related
to simple properties of individual sentences (e.g., in case when we
are testing if they have the same subject or direct object), we also
project sentence embeddings on a 2-D surface using UMAP
\citep{umap}\footnote{We use the default settings and pairwise cosine
  dissimilarities as distance measure.}  and check if
the spatial organisation of the points is in line with our
observations.


\paragraph{Lexical choice} A potential confound of our
experimental setup is lexical choice, which is never completely neutral.
For example, by taking a semantically close
pair of verbs, we can considerably reduce the effect of predicate
mismatch between two sentences. Moreover, encoders can react 
idiosyncratically to particular words and word combinations. Including all
combinations of words and their positions in sentence pairs 
as predictor variables is not a solution, however, as it defeats the purpose of
identifying structural patterns and, in the limit, amounts to replicating the
encoders. We address this confound in three ways.

First, we select nouns to be always at least as
interchangeable as words of other parts of speech in terms of belonging to similar mid-to-high frequency bands and referring to conceptually simple, concrete objects. This follows from our working
hypothesis that encoders give preferential treatment to nominal
elements, whose (generally entity referring) semantics is arguably
easier to capture than, for example, that of (generally event referring)
verbs \cite{baroni-lenci-2011-blessed}.


Second, we compare the analysis of the ST encoders against the analysis
of the vanilla BERT encoder. As they are derived from averaging,
vanilla BERT embeddings treat all words equally, so if our sentences, 
e.g., undersell differences in
adverbs because we chose two nearly synonymous ones, this should be visible
in the small coefficient tracking the impact of adverbs in the
regression model based on BERT embeddings. As will be shown below,
however, the hierarchy of coefficients for regression models of STs is
very different from that for vanilla BERT, which arguably indicates that
the role of lexical effects is minor.

Third, we re-run all reported models on sentences of the same
structure with different lexical content; see the Appendix for
details. We observe high stability of coefficients across
replications, higher for STs than for vanilla BERT. This further
corroborates the validity of our generalisations.

\section{Case Studies}\label{sec:case-studies}

This section presents a series of case studies testing the sensitivity
of embeddings produced by sentence transformers and BERT token averages to
properties of input sentences. We start with analysing simple intransitive sentences
(\S~\ref{ssec:intransitive}) and simple transitive sentences
(\S~\ref{ssec:transitive}). We then make specific aspects of
the structure more complex, analysing the effect of lengthy NPs
(\S~\ref{sssec:long-modifiers}) and coordinated VPs
(\S~\ref{ssec:coordinated-vps}). Finally, we look more closely at the
syntax-semantics interface by inverting the prototypical alignment of
POS tags and syntactic functions (predicative nominals and gerund
subjects, \S~\ref{ssec:predicative-nominals}) and by testing the
degree to which encoders track particular syntactic functions of verb
arguments (\S~\ref{ssec:participant-sets}).

\subsection{Simple Intransitive Sentences}\label{ssec:intransitive}

\paragraph{Data}
The main goal of the analysis of simple intransitive sentences 
is to check the relative contribution of their components
to their embeddings. We study a nearly-minimal sentence template
with a nominal subject, an adverbial adjunct, and an intransitive verb.
We construct a set of 256 sentences of the form 
\enquote{[det] [subj] [adverb] [verb][punct]}, 
where \texttt{det} ranges over \{\textit{a}, \textit{the}\}; \texttt{subj} 
ranges over a set of nouns,\footnote{\{\textit{cat}, \textit{dog}, \textit{artist}, 
\textit{teacher}, \textit{planet}, \textit{star}, \textit{wind}, \textit{rain}\}}
\texttt{adverb} ranges over \{\textit{quickly}, \textit{slowly}\}, \texttt{verb} ranges over
\{\textit{appears}, \textit{vanishes}, \textit{stops}, \textit{moves}\},
and \texttt{punct}, over \{\textit{.}, \textit{!}\}. Here and in subsequent experiments,
the generation procedure assures that all sentence features are statistically
independent, which is a crucial prerequisite for linear-regression modelling.

\paragraph{Model}
The regression model matrix is based on 32,640 pairs of generated sentences,
which differ in the value of at least one feature, with
predictor variables SameDeterminer, SameAdverb, SameVerb, SamePunct, and SameSubj.
We regress z-score-transformed cosine similarities between
sentence embeddings computed by three different encoders on these
predictor variables. The coefficients of the fitted models are shown in 
Table~\ref{tab:intransitives}.\footnote{Replication models,
fitted on sentences with the same structure but different lexical content, are
shown in Table~\ref{tab:app-intransitives} in the Appendix.}

\begin{table}[t]
\centering
\begin{tabular}{@{}lrrr@{}}
\toprule
 & \textbf{mpnet} & \textbf{distilroberta} & \textbf{bert} \\ \midrule
\textbf{SameDet} & 0.07 & 0.07 & 0.37 \\
\textbf{SameAdv} & 0.33 & 0.31 & 0.45 \\
\textbf{SamePred} & 0.74 & 0.61 & 0.58 \\
\textbf{SamePunct} & 0.24 & 0.24 & 0.84 \\
\textbf{SameSubj} & 2.26 & 2.40 & 1.27 \\ \midrule
\textbf{R-squared} & 0.67 & 0.71 & 0.48 \\ \bottomrule
\end{tabular}
\caption{A summary of the models predicting z-scored pairwise cosine similarities between
embeddings of sentences with intransitive verbs. All coefficients are significant with $p <$ 0.001.}\label{tab:intransitives}
\end{table}

\paragraph{Results}
Three observations from Table~\ref{tab:intransitives} hold for all subsequent analyses.

(i)~The coefficients are positive for all models and all features. This means that sentence pairs which agree in some constituent are always more similar than sentence pairs that do not -- as expected.

(ii)~The coefficient of determination ($R^2$) is larger for ST-focused linear models.
This means that the embeddings computed by the ST encoders are more dependent on the 
features of the sentences we track and less dependent on identities 
of lexical units. (It can be noted that the fact that we achieve \(R^2 \approx 0.7\)
using only a few structural properties is remarkable in itself.)

(iii)~The differences among coefficients of the ST-focused linear models are
in general larger than those of the linear model analysing BERT:
in the latter, the biggest coefficient (1.27 for SameSubj) is only $\approx$ 3.5 times 
higher than the smallest one (0.37 for SameDet), while for the ST models this
ratio is above 30. This is connected to the fact that BERT-derived sentence representations are 
more dependent on semantically impoverished elements, such as determiners and punctuation
signs, which dampen the effect of other constituents.
For the sake of brevity, we do not
analyse determiners and punctuation in subsequent experiments and
keep them constant as \textit{the} and \textit{.} respectively.

Turning to the comparison of coefficients inside models, we see that
STs pay considerably more attention to subjects than to predicates:
all things being equal, sentences with different predicates and
adverbs but the same subject will be more similar than sentences with
the same predicate and adverb and different subjects. The influence of
punctuation is surprisingly strong, being comparable to that of
adverbs, while the effect of determiners is very weak, albeit
statistically significant.

\begin{figure}[t]
    \centering
    \includegraphics[width=.5\textwidth]{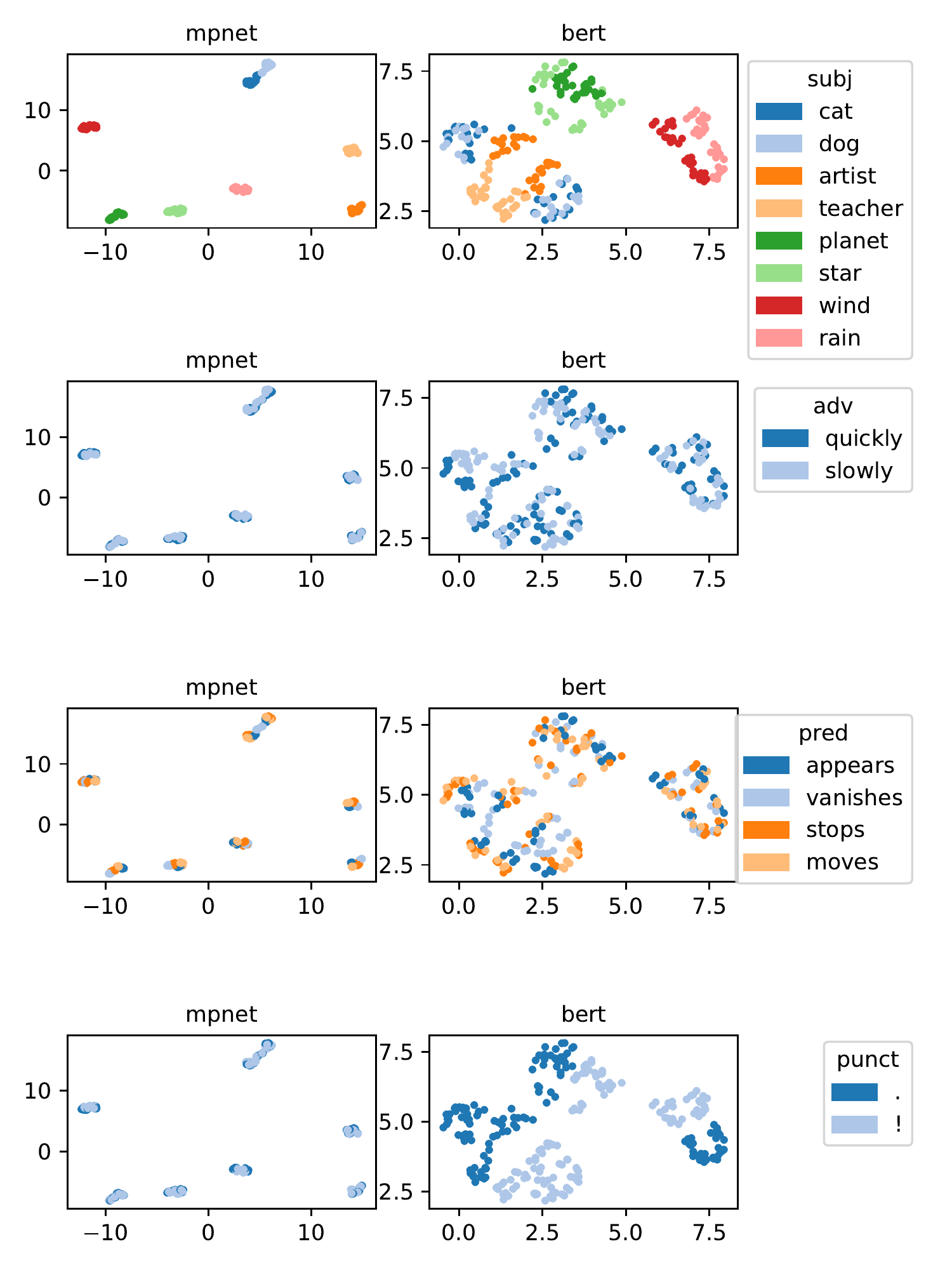}
    \caption{UMAP projections of embeddings of sentences with intransitive verbs (left: sentence transformer, right: BERT).}    \label{fig:intransitives-small}
\end{figure}

A plot of UMAP projections of sentence embeddings produced by MPNET and BERT, shown
in Figure~\ref{fig:intransitives-small}, underlines that while averaged BERT embeddings
distinguish punctuation signs but do not distinguish subjects, the situation is reversed 
for the sentence transformer: it distinguishes subjects cleanly but largely abstracts away 
from other structural properties.


\subsection{Transitive Sentences}\label{ssec:transitive}


\paragraph{Data}
The transitive sentences used in the analysis are generated using the following template:
\enquote{The [subj] [adverb] [verb] the [obj].}
The range of nouns was slightly extended;\footnote{To \{\textit{cat}, \textit{dog}, \textit{teacher}, 
\textit{artist}, \textit{robot}, \textit{machine}, \textit{tree}, \textit{bush}, \textit{planet}, 
\textit{star}, \textit{wind}, \textit{rain}\}.} the same adverbs as in the previous experiment were used, 
while \texttt{verb} ranged over \{\textit{sees}, \textit{chases}, \textit{draws}, \textit{meets}, 
\textit{remembers}, \textit{pokes}\}. This produces 672 different sentences and 225,456 sentence pairs.

\paragraph{Model}
The coding for SameAdv and SamePred remains as above. The main focus in this study is on whether
sentence similarities are dominated by the sentences having the same subject, the same direct object,
or the same words in these two positions even if their order were reversed. To test for this, we added
a categorical variable with the following values:

\begin{compactitem}
\item[\textbf{00}] no overlap in subject and object (the baseline);
\item[\textbf{A0}] same subject, different objects;
\item[\textbf{0B}] same object, different subjects;
\item[\textbf{0A}] the subject of the first sentence is the object of the second;
\item[\textbf{B0}] the object of the first sentence is the subject of the second;
\item[\textbf{BA}] subject and object are swapped;
\item[\textbf{AB}] the same subject and object.  
\end{compactitem}

\paragraph{Results}
A summary of the fitted models is given in Table~\ref{tab:transitives}.\footnote{A summary
of the replication model fits is provided in Table~\ref{tab:app-transitives} in the
Appendix.} It
demonstrates that when it comes to simple transitive sentences, our understanding
of their embeddings produced by sentence transformers remains high, despite
the sentences being more complex (\(R^2 \approx 0.7\)),
while BERT embeddings become more unpredictable (\(R^2 \approx 0.31\)).
Furthermore, while BERT again essentially treats all tokens more or less
equally, with adverbs slightly discounted, STs prioritise participants
(even B0 has higher coefficients than SamePred).

On the other hand, neither BERT nor STs prioritise the exact syntactic
function of the participants: coefficients for A0 vs.\ 0A, 0B vs.\ B0,
and AB vs.\ BA are largely comparable across all models with
\(BA \approx A0 + 0B\). That is, the effects of subjects and objects
are largely independent of one another.

A UMAP plot with the embeddings for the transitive sentences
is shown in Figure~\ref{fig:transitives} in the Appendix. It demonstrates that
STs arrive at a much more fine-grained clustering of sentences,
largely dominated by subjects and objects. They largely discount predicates and adverbs which are
quite prominent in averaged BERT embeddings.

\begin{table}[t]
\centering
\begin{tabular}{@{}lrrr@{}}
\toprule
 & \textbf{mpnet} & \textbf{distilroberta} & \textbf{bert} \\ \midrule
\textbf{SameAdv} & 0.49 & 0.36 & 0.56 \\
\textbf{SamePred} & 0.73 & 0.42 & 0.78 \\
\textbf{SubjObj\_0A} & 1.27 & 1.40 & 0.65 \\
\textbf{SubjObj\_0B} & 1.31 & 1.45 & 0.69 \\
\textbf{SubjObj\_A0} & 1.44 & 1.45 & 0.75 \\
\textbf{SubjObj\_AB} & 2.98 & 3.08 & 1.60 \\
\textbf{SubjObj\_B0} & 1.37 & 1.42 & 0.58 \\
\textbf{SubjObj\_BA} & 2.85 & 2.98 & 1.39 \\ \midrule
\textbf{R-squared} & 0.74 & 0.73 & 0.31 \\ \bottomrule
\end{tabular}
\caption{A summary of the models predicting z-scored pairwise cosine similarities between
embeddings of sentences with transitive verbs. All coefficients are significant with $p <$ 0.001.}\label{tab:transitives}
\end{table}

\subsection{Transitive Sentences with Long NP Modifiers}\label{sssec:long-modifiers}

The previous analyses showed that representations computed by STs are
highly attuned to verb participants but not to their particular
syntactic roles. This may mean that ST may be potentially misled by
nouns in other positions in the sentence, which have less relevance to
the described situation. This study explores this possibility.

\paragraph{Data} We repeat the analysis from
\S~\ref{ssec:transitive} using the template of the form
\enquote{The [subj] [modifier] [adverb] [verb] the [obj]}, with a
smaller set of subjects,\footnote{\{\textit{cat}, \textit{dog},
  \textit{rat}, \textit{giraffe}, \textit{wombat}, \textit{hippo}\}}
and the \texttt{modifier} ranging over \{\textit{with big shiny eyes},
\textit{that my brother saw yesterday}, \textit{whose photo was in the
  papers}, \textit{worth a great deal of money}\}.  Altogether this
gives 1,440 sentences and 1,036,080 sentence pairs.  The modifiers have
internal syntactic structure and contain a non-negligible amount of lexical
material that the models have to \enquote{skip over} if their
representations were focused on the participant structure of the matrix
clause.

\begin{table}[t]
\centering
\begin{tabular}{@{}lrrr@{}}
\toprule
 & \textbf{mpnet} & \textbf{distilroberta} & \textbf{bert} \\ \midrule
\textbf{SameMod} & 1.01 & 1.02 & \textbf{1.62} \\
\textbf{SameAdv} & 0.40 & 0.42 & 0.27 \\
\textbf{SamePred} & 0.89 & 0.67 & 0.40 \\
\textbf{SubjObj\_0A} & 0.83 & 1.06 & 0.32 \\
\textbf{SubjObj\_0B} & 0.97 & 1.27 & 0.42 \\
\textbf{SubjObj\_A0} & 1.11 & 1.14 & 0.53 \\
\textbf{SubjObj\_AB} & \textbf{2.14} & \textbf{2.44} & 1.00 \\
\textbf{SubjObj\_B0} & 1.20 & 1.30 & 0.54 \\
\textbf{SubjObj\_BA} & \textbf{2.09} & \textbf{2.40} & 0.91 \\ \midrule
\textbf{R-squared} & 0.73 & 0.81 & 0.61 \\ \bottomrule
\end{tabular}
\caption{A summary of the models predicting z-scored pairwise cosine similarities between
embeddings of sentences with transitive verbs and lengthy subject modifiers. All coefficients are significant with $p <$ 
0.001.}\label{tab:transitives-w-modifiers}
\end{table}

\paragraph{Model} The same coding strategy as in the preceding section
is used, augmented by a new binary variable, SameMod, tracking
whether two sentences have the same modifier for the subject.

\paragraph{Results} Both the model coefficients, shown in
Table~\ref{tab:transitives-w-modifiers}, and the UMAP plot, shown in
Figure~\ref{fig:transitives-w-modifiers} in the Appendix, indicate that
BERT embeddings are highly sensitive to lengthy modifiers:\footnote{The results
of the replication fits are shown in Table~\ref{tab:app-transitives-w-modifiers}
in the Appendix.} the SameMod coefficient in the linear model is larger
than the coefficients for the same predicate and the same subject-object
combination added together. The
situation is very different for STs: SameMod is more
important than SamePred, especially for \DR, but,
with one exception, not more important than even a partial overlap in
participants. Having the same participants, in either the same or
swapped syntactic functions, is more than twice as important. We take this as evidence  that STs have a specific bias towards matrix-clause
\textit{participant sets}, that is, the nouns that fill a thematic
role of the main predicate, while their precise functions and nouns
found in other positions in the sentence are less important.

\subsection{Coordinated Verbal Phrases}\label{ssec:coordinated-vps}

The analyses presented above show that the main predicate of the sentence has
only a limited influence on the representations computed by STs, compared
to its subjects and objects. Here, we show that this effect still holds
if there is more than one main predicate.

\paragraph{Data} Using the same sets of nouns and transitive verbs as
in the previous experiment, we construct sentences of the form
\enquote{The man [verb1] the [noun1], [verb2] the [noun2], and [verb3]
  the [noun3]}, where triples of verbs and nouns are taken from the
Cartesian product of the sets of all noun and verb combinations of size 3 
without replacement. To alleviate a possible ordering bias, all verb and 
noun triples are shuffled for each sentence.  This results in 400
sentences and 79,800 sentence pairs.

\paragraph{Models and results} The analysis proceeds in three
stages. First, we check if positions 1, 2, and 3 have different
importance by regressing the normalised cosine similarity on six
binary variables N[oun]1Same, V[erb]1Same, N2Same, etc.
The models, summarised in
Table~\ref{tab:coordinated-vp-dumb},\footnote{A summary of the
  replication fits is given in Table~\ref{tab:app-coordinated-vp-dumb}
  in the Appendix.} show low coefficients of determination (with
$R^2$ around 0.1), but they indicate that positions are of
unequal importance: BERT gives more weight to the last noun and the
last verb, while STs focus on the first and the last N-V pair
and largely ignore the second one.

\begin{table}[t]
\centering
\begin{tabular}{@{}lrrr@{}}
\toprule
                   & \textbf{mpnet} & \textbf{distilroberta} & \textbf{bert} \\ \midrule
\textbf{V1Same} & 0.41 & 0.26 & 0.21 \\
\textbf{V2Same} & 0.13 & 0.08 & 0.23 \\
\textbf{V3Same} & 0.36 & 0.34 & \textbf{0.41} \\
\textbf{N1Same} & 0.33 & 0.35 & 0.23 \\
\textbf{N2Same} & 0.12 & 0.22 & 0.30 \\
\textbf{N3Same} & \textbf{0.56} & \textbf{0.57} & \textbf{0.41} \\ \midrule
\textbf{R-squared} & 0.11 & 0.1 & 0.09 \\ \bottomrule
\end{tabular}
\caption{A summary of the models predicting z-scored pairwise cosine similarities between
embeddings of sentences with coordinated VPs from binary predictors.
All coefficients are significant with $p <$ 0.001.}\label{tab:coordinated-vp-dumb}
\end{table}

A significantly better fit can be achieved by replacing binary
predictors with overlap scores for nouns and verbs. As
Table~\ref{tab:coordinated-vp-overlap}\footnote{See
  Table~\ref{tab:app-coordinated-vp-overlap} in the Appendix for the
  replication fits.} shows, this type of model, even though it
contains only 2 variables instead of 6, obtains
$R^2 \approx 0.65$ for STs. It is also evident that all three models
place more weight on noun overlap than on verb overlap, with
\DR\ showing the biggest difference between the two.

This raises the question of whether particular verb-noun collocations
play a noticeable role, i.e., if a sentence containing
\textit{chases the wombat} will be considerably more similar to another
sentence containing the exact phrase compared to a sentence containing 
\textit{chases} and \textit{wombat} but not as a trigram. Simply 
adding $n$-gram overlap scores to the model is not possible,
however, because it is highly correlated with both noun overlap and
verb overlap. In order to obviate this obstacle, we first construct
an auxiliary linear model predicting trigram overlap from noun and
verb overlap and then use the residuals of this regression in the
main model.

The results are ambiguous: on one hand, the coefficient for
residualised trigram overlap is statistically significant with
$p < 0.001$. On the other hand, the effect is very weak (more than ten
times weaker than that of either noun overlap or verb overlap), and
the addition of trigram overlap to the model improves $R^2$ by less
than 0.001. This seems to indicate that trigram overlap is not
important for practical purposes.

\begin{table}[t]
\centering
\begin{tabular}{@{}lrrr@{}}
\toprule
                     & \textbf{mpnet} & \textbf{distilroberta} & \textbf{bert} \\ \midrule
\textbf{VerbOverlap} & 0.78           & 0.59                   & 0.64                \\
\textbf{NounOverlap} & 0.93           & 1.09                   & 0.88                \\ \midrule
\textbf{R-squared}   & 0.65           & 0.68                   & 0.52                \\ \bottomrule
\end{tabular}
\caption{A summary of the models predicting z-scored pairwise cosine similarities between
embeddings of sentences with coordinated VPs from overlap scores. All coefficients are significant with 
$p < 0.001$.}\label{tab:coordinated-vp-overlap}
\end{table}

\subsection{Predicative Nominals with Gerund Subjects}\label{ssec:predicative-nominals}

A potential weak point of our analysis is that parts of speech and
syntactic functions are not decoupled: it is not yet clear whether
the encoders pay attention to nouns or to subjects and objects.

\paragraph{Data} To address this issue, we construct another set of
sentences where the subject is a gerund and the predicate is
nominal. The template is \enquote{[gerund] [object] [copula] a
  [adjective] [predicate]}, where \texttt{gerund} ranges over
\{\textit{continuing}, \textit{abandoning}, \textit{starting},
\textit{completing}\}, object ranges over \{\textit{it},
\textit{them}, \textit{the project}, \textit{the plan}\}, copula is
one of \{\textit{is}, \textit{was}, \textit{will be}, \textit{is going
  to be}\}, adjectives are \{\textit{big}, \textit{real},
\textit{negligible}, \textit{insignificant}\}, and the predicative
nominal ranges over \{\textit{solution}, \textit{mistake},
\textit{failure}, \textit{triumph}\}. This gives 1024 sentences and
523,776 sentence pairs. A~variable copula provides an additional test
as to whether the sentence encoders can recognise multi-word sequences
with low semantic content.

\begin{table}[t]
\begin{tabular}{@{}lrrr@{}}
\toprule
                     & \textbf{mpnet} & \textbf{distilroberta} & \textbf{bert} \\ \midrule
\textbf{SameSubj}    & 0.82  & 0.70          & 0.31       \\
\textbf{SameCop}     & 0.35  & 0.30          & 0.55       \\
\textbf{SameAdj}     & 0.58  & 0.79          & 0.50       \\
\textbf{SamePred}    & 0.99   & 1.01          & 0.52       \\
\textbf{SameObjNoun} & 1.01  & 1.04          & 0.60       \\
\textbf{SameObjPron} & 0.44  & 0.50         & 0.42       \\ \midrule
\textbf{R-squared}   & 0.50  & 0.54          & 0.22       \\ \bottomrule
\end{tabular}
\caption{A summary of the models predicting z-scored pairwise cosine similarities between
embeddings of sentences with gerund subjects and nominal predicates.
All coefficients are significant with $p < 0.001$.}\label{tab:gerund-subjects}
\end{table}

\paragraph{Model} The sentence pair encoding includes four binary
variables (SameSubj, SameCop, SameAdj, SamePred) and a nominal
variable for the direct object, indicating whether objects are
different (baseline), are identical and pronominal (SamePron), or are
identical and nominal (SameNoun).

\paragraph{Results} The results in Table~\ref{tab:gerund-subjects}\footnote{See
an overview of replication fits in Table~\ref{tab:app-gerund-subjects}
in the Appendix.} demonstrate that all models treat both nominal predicates and nominal
direct objects as more important than gerund subjects. STs, moreover,
pay less attention to identical pronominal objects and discount
multi-word copula forms. \(R^2\) values for the ST model are lower
than in the previous experiments (in the 0.50--0.55 range), which may
potentially indicate a poor choice of lexical items; however,
replication experiments with a different set of words (except for
copula forms) achieved comparable results. This suggests that
embeddings of sentences of this type are less easily explainable as 
additive combinations of individual words compared to the sentence
types surveyed previously.

\subsection{Revisiting Participant Sets: Ditransitive Sentences}\label{ssec:participant-sets}

Our final experiment revisits the opposition between lexical overlap
in verbal phrases and exact argument-predicate matching. In this case,
we focus on ditransitive verbs with two arguments: a direct object and
an oblique object which is an integral part of the
situation.\footnote{Many English ditransitive verbs can
  undergo the \enquote{dative alternation}, which swaps the oblique object with
  a prepositional phrase: \textit{Give the book to me/John} vs.\
  \textit{Give me/John a book} \citep{levin1993english}. Of the verbs we
  use, \textit{show} and \textit{sell} participate in it, and the status of 
  \textit{describe} varies across speakers.}

\paragraph{Data}
All permutations of the triple of basic nouns \{\textit{cat},
\textit{dog}, \textit{rat}\} are generated.  For each permutation, all
three nouns are, in turn, replaced with one of the members of the set
of extra nouns \{\textit{giraffe}, \textit{wombat}, \textit{hippo}\};
the original permutations are also used. This provides a set of unique
triples of nouns where each pair of triples has from one to three
nouns in common. The Cartesian product of this set of triples with a
set of ditransitive verbs (\{\textit{describes}, \textit{sells},
\textit{shows}\}) and a set of adverbs (\{\textit{happily},
\textit{quickly}, \textit{secretly}) is used to fill the template
\enquote{The [noun1] [adverb] [verb] the [noun2] to the [noun3].} This
procedures gives 540 sentences and 145,530 sentence pairs.

\begin{table}[t]
\centering
\begin{tabular}{@{}lrrr@{}}
\toprule
 & \textbf{mpnet} & \textbf{distilroberta} & \textbf{bert} \\ \midrule
\textbf{SameAdv} & 1.05 & 1.07 & 0.64 \\
\textbf{SamePred} & 0.93 & 0.64 & 0.83 \\
\textbf{Overlap} & 0.90 & 1.00 & 0.91 \\
\textbf{SPCRes} & 0.03 & 0.02 & 0.10 \\ \midrule
\textbf{R-squared} & 0.745 & 0.738 & 0.57 \\ 
\textbf{\begin{tabular}[c]{@{}l@{}}R-squared\\ (w/o SPCRes)\end{tabular}} & 0.744 & 0.737 & 0.56 \\ \bottomrule
\end{tabular}
\caption{A summary of the models predicting z-scored pairwise cosine similarities between
embeddings of sentences with ditransitive verbs. SPCRes stands for SamePosCountRes, i.e.\ the
residuals of the number of identical words in identical positions regressed on lexical overlap.
All coefficients are significant with $p < 0.001$.}\label{tab:ditransitives}
\end{table}

\paragraph{Model}
The sentence pairs are coded for same adverb, same predicate, the
number of matching nouns in matching positions (SamePosCount), and
lexical overlap minus 1 (the baseline value of 0 corresponds to
overlap of 1; each successive value corresponds to increase in
overlap). As with overlapping words and trigrams above, these
predictors are correlated. Therefore, we residualise SamePosCount
after regressing it on lexical overlap.

\paragraph{Results} Table~\ref{tab:ditransitives} is
inconclusive in a similar way to results from
\S~\ref{ssec:predicative-nominals}. The coefficients for
residualised SamePosCount are significant; however, in the ST models, 
their size is very small, and SamePosCount does not materially improve 
the predictive power. We conclude, therefore, that
syntactic positions do not matter a great deal, in line with
our \enquote{participant set} interpretation from~\S~\ref{ssec:coordinated-vps}.

\section{Discussion}\label{sec:discussion}

Our analysis arguably goes some way towards explaining why sentence
transformers beat vanilla BERT-based models with token averaging on
sentence-modelling tasks.  Token averaging makes it impossible to
distinguish between semantically rich and impoverished sentence
elements, nor between syntactically central vs.\ peripheral elements:
punctuation signs and determiners contribute on the same level as
the matrix-clause predicate and main participants, while lengthy
modifiers, such as relative clauses, and multi-word copula forms
dominate the representation.

Sentence transformers, on the other hand, learn to discount elements
that only serve a grammatical function or present background
information and focus instead on the semantic kernel of the
sentence. The latter is in effect largely synonymous with the set of
nominal elements in the main clause, first of all participants, but
also predicative nominals. Importantly, despite their evident
syntactic-analytic capabilities (e.g., in our setting they can
distinguish between participants of main and relatives clauses
and between main and auxiliary verbs), STs seem to not pay much
attention to the distinction between subjects and direct or indirect
objects. Instead they prioritise raw overlap in the set of nominal participants
of the matrix clause. This can be seen, by slightly abusing
terminology of theoretical linguistics, as a focus on the
aboutness/topic of sentences, what things they describe, and not on
their predication/comment, what they actually say about those things
\citep{hu2009decomposing}.

We believe that this focus is not inherent to the architecture of
sentence transformers but reflects the nature of the datasets used
for fine-tuning STs. The size of these datasets makes it impossible to
convincingly reason about their contents, but their genres (QA pairs,
Reddit threads, etc.) makes it plausible to expect a high degree of topic-based overlap: questions
and conversations tend to revolve around entities (persons and
things), with their actions and properties repeating less often. This
naturally leads to a focus on nouns referring to prominent entities,
which are known to appear preferentially as subjects or objects for
reasons of coherence \cite{barzilay-lapata-2008-modeling}, arguably a
good match to the patterns we observe.


\section{Related Work}\label{sec:related-work}

Analysis of transformer-based models for sentence-level tasks, such
as NLI, question answering, or text classification, has largely followed
the same approaches as found in the general BERTology
\citep{rogers-etal-2020-primer}: probing, analysis of the geometry of
the embedding space, extraction of parts of input that are
particularly important for model performance, and behavioural analysis.
In this vein, \citet{liu2021exploring} and \citet{peyrard2021laughing}
analyse the attention patterns powering the performance of transformer
models on different types of sentence classification, and
\citet{li-etal-2020-sentence} show that embeddings of sentences
computed by BERT-based models, including siamese-fine-tuned sentence
transformers, are anisotropic and can be improved via normalisation.
\citet{chrysostomou2021variable} survey the existing methods for
extracting rationales from input sentences in the context of text
classification and propose an improved approach, while
\citet{luo-etal-2021-positional} demonstrate that 
sentence embeddings derived by averaging BERT token representations
suffer from artefacts arising from positional embeddings. 
\citet{zhelezniak2019max} argue that averaging should be replaced with
max-pooling. 


Very similar to ours is the approach adopted by
\citet{macavaney2022abnirml}, who construct a series of probes to
analyse the performance of several models on the task of information
retrieval. While their methodology relies on high-level document
statistics and wholistic document manipulation (word and sentence
shuffling, token-frequency similarity between the document and the
query, textual fluency, etc.), our study analyses the role of
linguistically motivated structural factors and thus complements their
findings.

\citet{opitz2022sbert} aim at directly decomposing the representations 
produced by sentence transformers into several parts capturing different 
properties of sentences reflected in AMR annotations (presence of negation, 
concepts included in the sentence, etc.). While our study tries to
ascertain what meaning components dominate the representations, \citeauthor{opitz2022sbert}
assume that these components are known in advance and are equally important:
sentence embeddings in their modified SBERT model are split into 15 segments, 
each of which corresponds to one AMR-based meaning component,
plus a residual part to capture everything not
covered by AMR annotations.

\section{Conclusion}\label{sec:conclusion}


This paper aims at making a contribution towards a better
understanding of sentence transformers, which are often seen as black
boxes. We have demonstrated that we can make surprisingly precise
inferences about sentence-pair similarities using simple linguistic
features such as lexical overlap.

The crucial difference between bag-of-words distributional models 
and current encoders is that STs have became quite adept at disregarding
\enquote{irrelevant} parts of the
sentence and concentrating on its key elements. Unlike vanilla BERT sentence
embeddings obtained by token averaging, STs yield more structured embeddings
that focus on the matrix clause and are less tied to individual lexical
items and strings of function words.

This progress, however, comes with a particular type of bias:
the structures that lead to high sentence similarity in STs, 
i.e.\ the overlap in nominal \enquote{participant sets}, seem to mirror 
the dominant type of paraphrases found in the data the STs were tuned on, 
and STs are not compelled to look at finer structures of input sentences.
At least without further fine tuning, this would appear to make them 
unsuitable for downstream tasks that require knowledge about more fine-grained 
aspects of sentence structure, such as semantic roles \cite{conia-navigli-2022-probing},
or extra-propositional aspects, such as monotonicity, negation, or 
modality~\cite{yanaka-etal-2021-sygns,nakov-2016-negation}.

An interesting direction for future research would be to explore the ways of 
decomposing sentence representations into
additive aspects such as participant structure, main predication, etc.
The additional challenge here is that while theoretical semantics has a
lot to say about aspects of sentence meaning \citep{pagin2016semantics},
there remains a lack of analysis linking the notion of one-dimensional
\textit{semantic similarity} \cite{agirre-etal-2012-semeval} that underlies the 
optimisation of current sentence transformers with theoretically more substantial concepts.

\section*{Limitations}

The limitations of the proposed analysis are the following:

\begin{enumerate}
    \item The analysis is based on synthetic data. This allows us to fully control the sentence structure and use balanced lexical material, but it does not necessarily reflect the performance of models on real-world data, especially when sentences or text fragments are much longer. However, synthetic data have generally shown to be a good first step toward understanding the behaviour of complex models.
    \item The analysis does not cover graded distinctions between words, i.e.\ we did not experiment with filling the slots with synonymous words, as opposed to completely unrelated words. This makes it impossible to decide if the models are sensitive to word identities or to their actual semantics, as long as these two notions are distinguishable.
    \item The outputs of the models are interpreted using linear regression analysis anchored to the properties of synthetic sentences. This kind of analysis makes it possible to disentangle additive effects of different components of sentence structure and provides statistical-significance estimates, while high $R^2$ values indicate that our findings have some validity. However, it cannot fully account for the lexical effects (which we tried to safeguard against by carefully selecting template fillers), non-linear effects, and hidden collinearity patterns (beyond those we addressed using residualised analysis).
    \item The range of models analysed in the paper is restricted. It covers some amount of variability (sentence transformers vs.\ vanilla BERT; two different variants of a base model for STs, one of them distilled), but other combinations of model architecture and training/fine-tuning regime can lead to different outcomes.
\end{enumerate}


\bibliography{anthology,custom}

\begin{thebibliography}{32}
\expandafter\ifx\csname natexlab\endcsname\relax\def\natexlab#1{#1}\fi

\bibitem[{Agirre et~al.(2012)Agirre, Cer, Diab, and
  Gonzalez-Agirre}]{agirre-etal-2012-semeval}
Eneko Agirre, Daniel Cer, Mona Diab, and Aitor Gonzalez-Agirre. 2012.
\newblock \href {https://aclanthology.org/S12-1051} {{S}em{E}val-2012 task 6: A
  pilot on semantic textual similarity}.
\newblock In \emph{*{SEM} 2012: The First Joint Conference on Lexical and
  Computational Semantics {--} Volume 1: Proceedings of the main conference and
  the shared task, and Volume 2: Proceedings of the Sixth International
  Workshop on Semantic Evaluation ({S}em{E}val 2012)}, pages 385--393,
  Montr{\'e}al, Canada. Association for Computational Linguistics.

\bibitem[{Baroni and Lenci(2011)}]{baroni-lenci-2011-blessed}
Marco Baroni and Alessandro Lenci. 2011.
\newblock \href {https://aclanthology.org/W11-2501} {How we {BLESS}ed
  distributional semantic evaluation}.
\newblock In \emph{Proceedings of the {GEMS} 2011 Workshop on {GE}ometrical
  Models of Natural Language Semantics}, pages 1--10, Edinburgh, UK.
  Association for Computational Linguistics.

\bibitem[{Barzilay and Lapata(2008)}]{barzilay-lapata-2008-modeling}
Regina Barzilay and Mirella Lapata. 2008.
\newblock \href {https://doi.org/10.1162/coli.2008.34.1.1} {Modeling local
  coherence: An entity-based approach}.
\newblock \emph{Computational Linguistics}, 34(1):1--34.

\bibitem[{Chrysostomou and Aletras(2021)}]{chrysostomou2021variable}
George Chrysostomou and Nikolaos Aletras. 2021.
\newblock Variable instance-level explainability for text classification.
\newblock \emph{arXiv preprint arXiv:2104.08219}.

\bibitem[{Conia and Navigli(2022)}]{conia-navigli-2022-probing}
Simone Conia and Roberto Navigli. 2022.
\newblock \href {https://doi.org/10.18653/v1/2022.acl-long.316} {Probing for
  predicate argument structures in pretrained language models}.
\newblock In \emph{Proceedings of the 60th Annual Meeting of the Association
  for Computational Linguistics (Volume 1: Long Papers)}, pages 4622--4632,
  Dublin, Ireland. Association for Computational Linguistics.

\bibitem[{Devlin et~al.(2019)Devlin, Chang, Lee, and
  Toutanova}]{devlin-etal-2019-bert}
Jacob Devlin, Ming-Wei Chang, Kenton Lee, and Kristina Toutanova. 2019.
\newblock \href {https://doi.org/10.18653/v1/N19-1423} {{BERT}: Pre-training of
  deep bidirectional transformers for language understanding}.
\newblock In \emph{Proceedings of the 2019 Conference of the North {A}merican
  Chapter of the Association for Computational Linguistics: Human Language
  Technologies, Volume 1 (Long and Short Papers)}, pages 4171--4186,
  Minneapolis, Minnesota. Association for Computational Linguistics.

\bibitem[{Ha et~al.(2021)Ha, Nguyen, Nguyen, Nguyen, and
  Than}]{ha2021utilizing}
Thi-Thanh Ha, Van-Nha Nguyen, Kiem-Hieu Nguyen, Kim-Anh Nguyen, and Quang-Khoat
  Than. 2021.
\newblock Utilizing {SBERT} for finding similar questions in community question
  answering.
\newblock In \emph{2021 13th International Conference on Knowledge and Systems
  Engineering (KSE)}, pages 1--6. IEEE.

\bibitem[{Hu and Pan(2009)}]{hu2009decomposing}
Jianhua Hu and Haihua Pan. 2009.
\newblock Decomposing the aboutness condition for {Chinese} topic
  constructions.
\newblock \emph{The Linguistic Review}, 26:371--384.

\bibitem[{Kiritchenko and Mohammad(2018)}]{kiritchenko-mohammad-2018-examining}
Svetlana Kiritchenko and Saif Mohammad. 2018.
\newblock \href {https://doi.org/10.18653/v1/S18-2005} {Examining gender and
  race bias in two hundred sentiment analysis systems}.
\newblock In \emph{Proceedings of the Seventh Joint Conference on Lexical and
  Computational Semantics}, pages 43--53, New Orleans, Louisiana. Association
  for Computational Linguistics.

\bibitem[{Levin(1993)}]{levin1993english}
Beth Levin. 1993.
\newblock \emph{English verb classes and alternations: {A} preliminary
  investigation}.
\newblock University of Chicago Press.

\bibitem[{Li et~al.(2020)Li, Zhou, He, Wang, Yang, and
  Li}]{li-etal-2020-sentence}
Bohan Li, Hao Zhou, Junxian He, Mingxuan Wang, Yiming Yang, and Lei Li. 2020.
\newblock \href {https://doi.org/10.18653/v1/2020.emnlp-main.733} {On the
  sentence embeddings from pre-trained language models}.
\newblock In \emph{Proceedings of the 2020 Conference on Empirical Methods in
  Natural Language Processing (EMNLP)}, pages 9119--9130, Online. Association
  for Computational Linguistics.

\bibitem[{Liu et~al.(2021)Liu, Le, Chakraborty, and
  Abdelzaher}]{liu2021exploring}
Shengzhong Liu, Franck Le, Supriyo Chakraborty, and Tarek Abdelzaher. 2021.
\newblock On exploring attention-based explanation for transformer models in
  text classification.
\newblock In \emph{2021 IEEE International Conference on Big Data (Big Data)},
  pages 1193--1203. IEEE.

\bibitem[{Liu et~al.(2019)Liu, Ott, Goyal, Du, Joshi, Chen, Levy, Lewis,
  Zettlemoyer, and Stoyanov}]{liu2019roberta}
Yinhan Liu, Myle Ott, Naman Goyal, Jingfei Du, Mandar Joshi, Danqi Chen, Omer
  Levy, Mike Lewis, Luke Zettlemoyer, and Veselin Stoyanov. 2019.
\newblock Roberta: A robustly optimized bert pretraining approach.
\newblock \emph{arXiv preprint arXiv:1907.11692}.

\bibitem[{Luo et~al.(2021)Luo, Kulmizev, and Mao}]{luo-etal-2021-positional}
Ziyang Luo, Artur Kulmizev, and Xiaoxi Mao. 2021.
\newblock \href {https://doi.org/10.18653/v1/2021.acl-long.413} {Positional
  artefacts propagate through masked language model embeddings}.
\newblock In \emph{Proceedings of the 59th Annual Meeting of the Association
  for Computational Linguistics and the 11th International Joint Conference on
  Natural Language Processing (Volume 1: Long Papers)}, pages 5312--5327,
  Online. Association for Computational Linguistics.

\bibitem[{Ma et~al.(2019)Ma, Wang, Ng, Nallapati, and Xiang}]{ma2019universal}
Xiaofei Ma, Zhiguo Wang, Patrick Ng, Ramesh Nallapati, and Bing Xiang. 2019.
\newblock Universal text representation from {BERT}: {A}n empirical study.
\newblock \emph{arXiv preprint arXiv:1910.07973}.

\bibitem[{MacAvaney et~al.(2022)MacAvaney, Feldman, Goharian, Downey, and
  Cohan}]{macavaney2022abnirml}
Sean MacAvaney, Sergey Feldman, Nazli Goharian, Doug Downey, and Arman Cohan.
  2022.
\newblock {ABNIRML}: {A}nalyzing the behavior of neural {IR} models.
\newblock \emph{Transactions of the Association for Computational Linguistics},
  10:224--239.

\bibitem[{Marvin and Linzen(2018)}]{marvin-linzen-2018-targeted}
Rebecca Marvin and Tal Linzen. 2018.
\newblock \href {https://doi.org/10.18653/v1/D18-1151} {Targeted syntactic
  evaluation of language models}.
\newblock In \emph{Proceedings of the 2018 Conference on Empirical Methods in
  Natural Language Processing}, pages 1192--1202, Brussels, Belgium.
  Association for Computational Linguistics.

\bibitem[{McInnes et~al.(2018)McInnes, Healy, and Melville}]{umap}
Leland McInnes, John Healy, and James Melville. 2018.
\newblock \href {https://doi.org/10.48550/ARXIV.1802.03426} {{UMAP}: Uniform
  manifold approximation and projection for dimension reduction}.
\newblock \emph{ArXiv}, abs/1802.03426.

\bibitem[{Minaee et~al.(2021)Minaee, Kalchbrenner, Cambria, Nikzad, Chenaghlu,
  and Gao}]{minaee2021deep}
Shervin Minaee, Nal Kalchbrenner, Erik Cambria, Narjes Nikzad, Meysam
  Chenaghlu, and Jianfeng Gao. 2021.
\newblock Deep learning--based text classification: A comprehensive review.
\newblock \emph{ACM Computing Surveys (CSUR)}, 54(3):1--40.

\bibitem[{Nakov(2016)}]{nakov-2016-negation}
Preslav Nakov. 2016.
\newblock \href {https://aclanthology.org/W16-5005} {Negation and modality in
  machine translation}.
\newblock In \emph{Proceedings of the Workshop on Extra-Propositional Aspects
  of Meaning in Computational Linguistics ({E}x{P}ro{M})}, page~41, Osaka,
  Japan. The COLING 2016 Organizing Committee.

\bibitem[{Opitz and Frank(2022)}]{opitz2022sbert}
Juri Opitz and Anette Frank. 2022.
\newblock \href {https://aclanthology.org/2022.aacl-main.48} {{SBERT} studies
  meaning representations: Decomposing sentence embeddings into explainable
  semantic features}.
\newblock In \emph{Proceedings of the 2nd Conference of the Asia-Pacific
  Chapter of the Association for Computational Linguistics and the 12th
  International Joint Conference on Natural Language Processing (Volume 1: Long
  Papers)}, pages 625--638, Online only. Association for Computational
  Linguistics.

\bibitem[{Pagin(2016)}]{pagin2016semantics}
Peter Pagin. 2016.
\newblock \href {https://doi.org/https://doi.org/10.1017/CBO9781139236157.004}
  {Sentential semantics}.
\newblock In Maria Aloni and Paul Dekker, editors, \emph{Cambridge Handbook of
  Formal Semantics}, Cambridge Handbooks in Language and Linguistics, pages
  65--105. Cambridge University Press.

\bibitem[{Peyrard et~al.(2021)Peyrard, Borges, Gligori{\'c}, and
  West}]{peyrard2021laughing}
Maxime Peyrard, Beatriz Borges, Kristina Gligori{\'c}, and Robert West. 2021.
\newblock Laughing heads: Can transformers detect what makes a sentence funny?
\newblock \emph{arXiv preprint arXiv:2105.09142}.

\bibitem[{Reimers and Gurevych(2019)}]{reimers-gurevych-2019-sentence}
Nils Reimers and Iryna Gurevych. 2019.
\newblock \href {https://doi.org/10.18653/v1/D19-1410} {Sentence-{BERT}:
  Sentence embeddings using {S}iamese {BERT}-networks}.
\newblock In \emph{Proceedings of the 2019 Conference on Empirical Methods in
  Natural Language Processing and the 9th International Joint Conference on
  Natural Language Processing (EMNLP-IJCNLP)}, pages 3982--3992, Hong Kong,
  China. Association for Computational Linguistics.

\bibitem[{Rogers et~al.(2020)Rogers, Kovaleva, and
  Rumshisky}]{rogers-etal-2020-primer}
Anna Rogers, Olga Kovaleva, and Anna Rumshisky. 2020.
\newblock \href {https://doi.org/10.1162/tacl_a_00349} {A primer in
  {BERT}ology: What we know about how {BERT} works}.
\newblock \emph{Transactions of the Association for Computational Linguistics},
  8:842--866.

\bibitem[{Sanh et~al.(2019)Sanh, Debut, Chaumond, and
  Wolf}]{Sanh2019DistilBERTAD}
Victor Sanh, Lysandre Debut, Julien Chaumond, and Thomas Wolf. 2019.
\newblock {DistilBERT}, a distilled version of {BERT}: smaller, faster, cheaper
  and lighter.
\newblock \emph{ArXiv}, abs/1910.01108.

\bibitem[{Santander-Cruz et~al.(2022)Santander-Cruz, Salazar-Colores,
  Paredes-Garc{\'\i}a, Guendulain-Arenas, and
  Tovar-Arriaga}]{santander2022semantic}
Yamanki Santander-Cruz, Sebasti{\'a}n Salazar-Colores, Wilfrido~Jacobo
  Paredes-Garc{\'\i}a, Humberto Guendulain-Arenas, and Sa{\'u}l Tovar-Arriaga.
  2022.
\newblock Semantic feature extraction using {SBERT} for dementia detection.
\newblock \emph{Brain Sciences}, 12(2):270.

\bibitem[{Song et~al.(2020)Song, Tan, Qin, Lu, and Liu}]{song2020mpnet}
Kaitao Song, Xu~Tan, Tao Qin, Jianfeng Lu, and Tie-Yan Liu. 2020.
\newblock \href
  {https://proceedings.neurips.cc/paper/2020/file/c3a690be93aa602ee2dc0ccab5b7b67e-Paper.pdf}
  {{MPNet}: Masked and permuted pre-training for language understanding}.
\newblock In \emph{Proceedings of NeurIPS}, pages 16857--16867.

\bibitem[{Tenney et~al.(2019)Tenney, Das, and Pavlick}]{tenney-etal-2019-bert}
Ian Tenney, Dipanjan Das, and Ellie Pavlick. 2019.
\newblock \href {https://doi.org/10.18653/v1/P19-1452} {{BERT} rediscovers the
  classical {NLP} pipeline}.
\newblock In \emph{Proceedings of the 57th Annual Meeting of the Association
  for Computational Linguistics}, pages 4593--4601, Florence, Italy.
  Association for Computational Linguistics.

\bibitem[{Wang et~al.(2021)Wang, Jiang, Bach, Wang, Huang, Huang, and
  Tu}]{wang2021ner}
Xinyu Wang, Yong Jiang, Nguyen Bach, Tao Wang, Zhongqiang Huang, Fei Huang, and
  Kewei Tu. 2021.
\newblock \href {http://arxiv.org/abs/2105.03654} {Improving named entity
  recognition by external context retrieving and cooperative learning}.
\newblock \emph{CoRR}, abs/2105.03654.

\bibitem[{Yanaka et~al.(2021)Yanaka, Mineshima, and
  Inui}]{yanaka-etal-2021-sygns}
Hitomi Yanaka, Koji Mineshima, and Kentaro Inui. 2021.
\newblock \href {https://doi.org/10.18653/v1/2021.findings-acl.10} {{S}y{GNS}:
  A systematic generalization testbed based on natural language semantics}.
\newblock In \emph{Findings of the Association for Computational Linguistics:
  ACL-IJCNLP 2021}, pages 103--119, Online. Association for Computational
  Linguistics.

\bibitem[{Zhelezniak et~al.(2019)Zhelezniak, Savkov, Shen, Moramarco, Flann,
  and Hammerla}]{zhelezniak2019max}
Vitalii Zhelezniak, Aleksandar Savkov, April Shen, Francesco Moramarco, Jack
  Flann, and Nils~Y. Hammerla. 2019.
\newblock \href {http://arxiv.org/abs/1904.13264} {Don't settle for average, go
  for the max: Fuzzy sets and max-pooled word vectors}.
\newblock \emph{CoRR}, abs/1904.13264.

\end{thebibliography}
\bibliographystyle{acl_natbib}

\clearpage

\appendix

\section{Appendix}\label{sec:appendix}

\subsection{Dimensionality-reduction plots}\label{ssec:app-dimensionality-reduction}

\subsubsection{Simple transitive sentences}\label{sssec:app-transitive-plots}

\begin{figure*}[t]
    \centering
    \includegraphics[width=\textwidth]{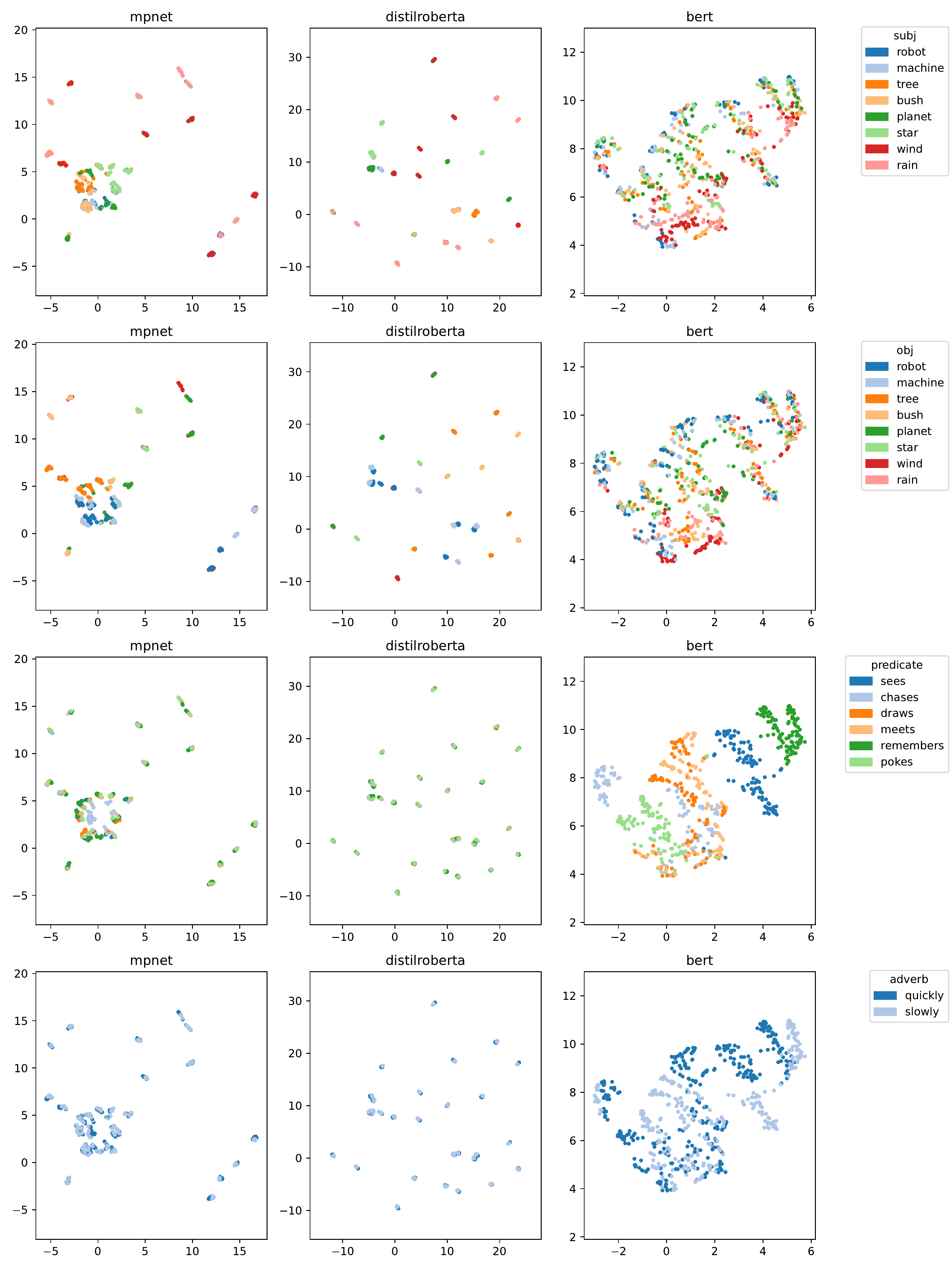}
    \caption{UMAP projections of embeddings of sentences with transitive verbs
    colour coded according to subject, object, predicate, and adverb.}
    \label{fig:transitives}
\end{figure*}

A UMAP plot of embeddings of simple transitive sentences encoded accordings to their
properties is shown in Figure~\ref{fig:transitives}.

\subsubsection{Transitive sentences with long NP modifiers}

\begin{figure*}[t]
    \centering
    \includegraphics[height=\textheight]{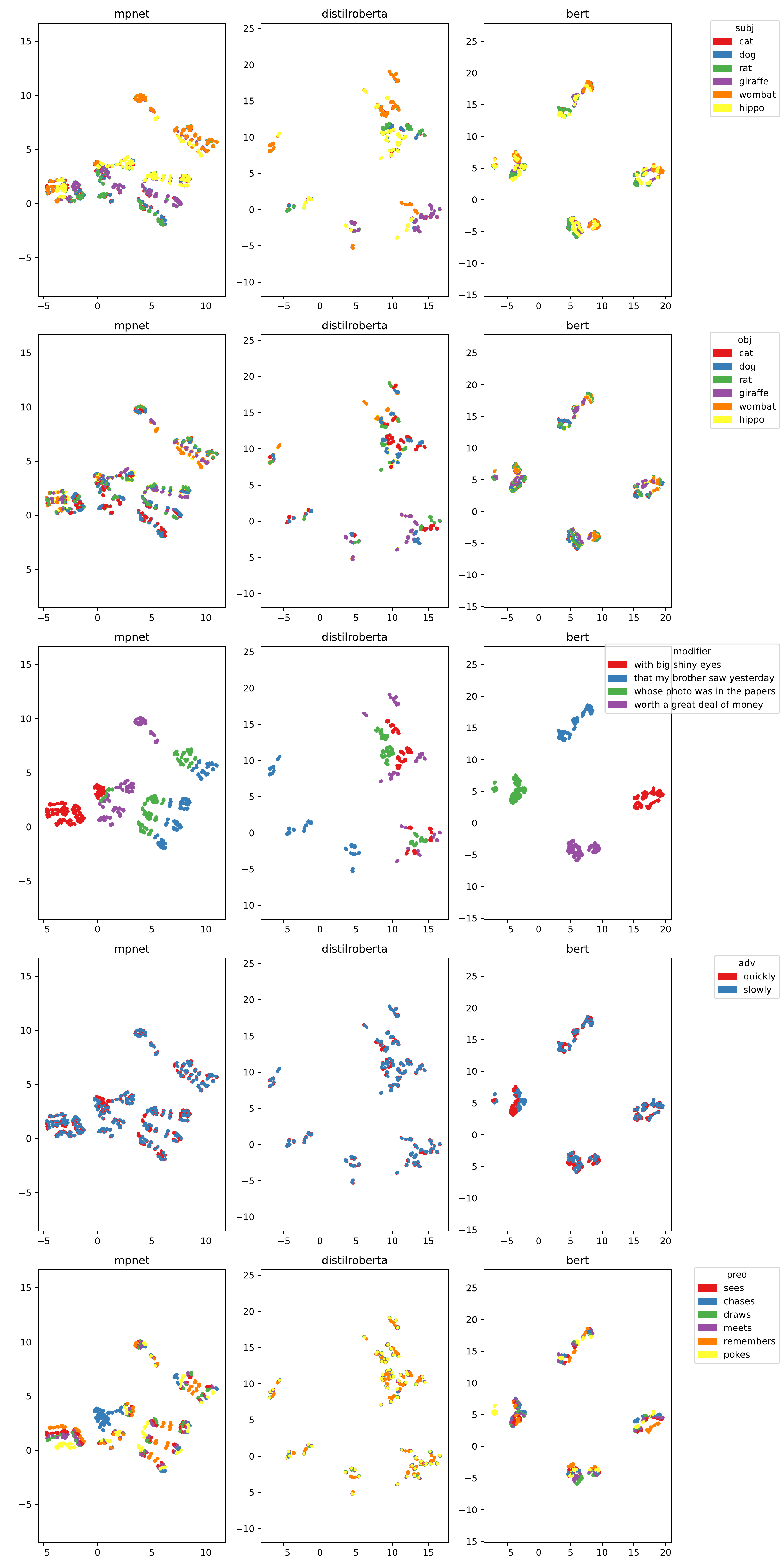}
    \caption{UMAP projections of embeddings of sentences with transitive verbs
    and long subject modifiers colour coded according to subject, modifier, object, predicate,
    and adverb.}
    \label{fig:transitives-w-modifiers}
\end{figure*}

A UMAP plot of embeddings of transitive sentences with lengthy subject modifiers
encoded accordings to their properties is shown in Figure~\ref{fig:transitives-w-modifiers}.

\subsection{Replication-model fits}\label{ssec:app-model-replications}

\subsubsection{Simple intransitive sentences}\label{sssec:app-simple-intransitives}

\begin{table}[t]
\centering
\begin{tabular}{@{}lrrr@{}}
\toprule
 & \textbf{mpnet} & \textbf{distilroberta} & \textbf{bert} \\ \midrule
\textbf{SameDet} & 0.08 & 0.11 & 0.26 \\
\textbf{SameAdv} & 0.38 & 0.38 & 0.96 \\
\textbf{SamePred} & 1.02 & 0.95 & 0.49 \\
\textbf{SamePunct} & 0.18 & 0.26 & 0.64 \\
\textbf{SameSubj} & 2.15 & 2.17 & 0.65 \\ \midrule
\textbf{R-squared} & 0.71 & 0.71 & 0.43 \\ \bottomrule
\end{tabular}
\caption{A summary of the replication models predicting z-scored pairwise cosine similarities between
embeddings of sentences with intransitive verbs. All coefficients are significant with $p <$ 0.001.}\label{tab:app-intransitives}
\end{table}

The following lexical items were used for the replication experiment:

\begin{itemize}
    \item Nouns: \textit{wolf}, \textit{bear}, \textit{fruit}, \textit{vegetable}, 
        \textit{building}, \textit{car}, \textit{lightning}, \textit{wave}
    \item Verbs: \textit{stabilizes}, \textit{bursts}, \textit{grows}, \textit{shrinks}
    \item Adverbs: \textit{suddenly}, \textit{predictably}
\end{itemize}
A summary of the replication models is shown in Table~\ref{tab:app-intransitives}.

\subsubsection{Simple transitive sentences}\label{sssec:app-simple-transitives}

\begin{table}[t]
\centering
\begin{tabular}{@{}lrrr@{}}
\toprule
 & \textbf{mpnet} & \textbf{distilroberta} & \textbf{bert} \\ \midrule
\textbf{SameAdv} & 0.54 & 0.32 & 0.95 \\
\textbf{SamePred} & 0.49 & 0.43 & 0.75 \\
\textbf{SubjObj\_0A} & 1.46 & 1.50 & 0.70 \\
\textbf{SubjObj\_0B} & 1.49 & 1.53 & 0.66 \\
\textbf{SubjObj\_A0} & 1.48 & 1.54 & 0.76 \\
\textbf{SubjObj\_AB} & 3.19 & 3.23 & 1.56 \\
\textbf{SubjObj\_B0} & 1.40 & 1.48 & 0.50 \\
\textbf{SubjObj\_BA} & 3.07 & 3.14 & 1.34 \\ \midrule
\textbf{R-squared} & 0.81 & 0.8 & 0.45 \\ \bottomrule
\end{tabular}
\caption{A summary of the replication models predicting z-scored pairwise cosine similarities between
embeddings of sentences with intransitive verbs. All coefficients are significant with $p <$ 0.001.}\label{tab:app-transitives}
\end{table}

The following lexical items were used for the replication experiment:

\begin{itemize}
    \item Nouns: \textit{pig}, \textit{horse}, \textit{soldier}, \textit{farmer}, \textit{android}, \textit{computer}, \textit{grass}, 
    \textit{forest}, \textit{comet}, \textit{galaxy}, \textit{cloud}, \textit{lightning}
    \item Verbs: \textit{hears}, \textit{pursues}, \textit{imagines}, \textit{recognizes}, \textit{touches}, \textit{finds}
    \item Adverbs: \textit{suddenly}, \textit{predictably}
\end{itemize}
A summary of the replication models is shown in Table~\ref{tab:app-transitives}.

\subsubsection{Transitive sentences with long NP modifiers}\label{sssec:app-long-modifiers}

The following lexical and phrasal items were used for the replication experiment:

\begin{itemize}
    \item Nouns: \textit{horse}, \textit{pig}, \textit{donkey}, \textit{elephant}, \textit{bison}, \textit{moose}
    \item NP modifiers: \textit{missing a hind leg}, \textit{whose face we all know}, \textit{born under a bad sign}, \textit{pictured on page seventeen}
    \item Verbs: \textit{hears}, \textit{pursues}, \textit{imagines}, \textit{recognizes}, \textit{touches}, \textit{finds}
    \item Adverbs: \textit{suddenly}, \textit{predictably}
\end{itemize}
The overview of the model fits is shown in Table~\ref{tab:app-transitives-w-modifiers}.

\begin{table}[t]
\centering
\begin{tabular}{@{}lrrr@{}}
\toprule
 & \textbf{mpnet} & \textbf{distilroberta} & \textbf{bert} \\ \midrule
\textbf{SameMod}     & 1.18          & 1.26          & \textbf{1.83} \\
\textbf{SameAdv}     & 0.48          & 0.26          & 0.41 \\
\textbf{SamePred}    & 0.64          & 0.64          & 0.44 \\
\textbf{SubjObj\_0A} & 0.91          & 1.00             & 0.18 \\
\textbf{SubjObj\_0B} & 0.99          & 1.09          & 0.17 \\
\textbf{SubjObj\_A0} & 1.10          & 1.19          & 0.24 \\
\textbf{SubjObj\_AB} & \textbf{2.13} & \textbf{2.32} & 0.42 \\
\textbf{SubjObj\_B0} & 1.16          & 1.25          & 0.20 \\
\textbf{SubjObj\_BA} & \textbf{2.11} & \textbf{2.28}  & 0.39 \\ \midrule
\textbf{R-squared}   & 0.77          & 0.84          & 0.71 \\ \bottomrule
\end{tabular}
\caption{A summary of the replication models predicting z-scored pairwise cosine similarities between
embeddings of sentences with transitive verbs and lengthy subject modifiers. All coefficients are significant with $p <$ 
0.001.}\label{tab:app-transitives-w-modifiers}
\end{table}

\subsubsection{Coordinated verbal phrases}\label{sssec:app-coordinated-vps}

The following lexical items were used for the replication experiment:

\begin{itemize}
    \item Nouns: \textit{mouse}, \textit{horse}, \textit{fox}, \textit{kangaroo}, \textit{bison}, \textit{elephant}
    \item Verbs: \textit{hears}, \textit{pursues}, \textit{imagines}, \textit{recognizes}, \textit{touches}, \textit{finds}
\end{itemize}
A summary of the replication models is shown in Tables~\ref{tab:app-coordinated-vp-dumb} 
(individual-word-based models)
and \ref{tab:app-coordinated-vp-overlap} (overlap-based models).

\begin{table}[t]
\centering
\begin{tabular}{@{}lrrr@{}}
\toprule
                   & \textbf{mpnet} & \textbf{distilroberta} & \textbf{bert} \\ \midrule
\textbf{V1Same} & 0.29 & 0.18 & 0.35 \\
\textbf{V2Same} & 0.13 & 0.08 & 0.28 \\
\textbf{V3Same} & 0.39 & 0.40 & \textbf{0.42} \\
\textbf{N1Same} & 0.49 & 0.48 & 0.14 \\
\textbf{N2Same} & 0.10 & 0.25 & 0.18 \\
\textbf{N3Same} & \textbf{0.57} & \textbf{0.52} & 0.17 \\ \midrule
\textbf{R-squared} & 0.12 & 0.11 & 0.07 \\ \bottomrule
\end{tabular}
\caption{A summary of the replication models predicting z-scored pairwise cosine
similarities between embeddings of sentences with coordinated VPs from binary predictors.
All coefficients are significant with $p <$ 0.001.}\label{tab:app-coordinated-vp-dumb}
\end{table}

\begin{table}[t]
\centering
\begin{tabular}{@{}lrrr@{}}
\toprule
                     & \textbf{mpnet} & \textbf{distilroberta} & \textbf{bert} \\ \midrule
\textbf{VerbOverlap} & 0.69           & 0.52                   & 0.85                \\
\textbf{NounOverlap} & 1.05           & 1.20                   & 0.47                \\ \midrule
\textbf{R-squared}   & 0.69           & 0.76                   & 0.41                \\ \bottomrule
\end{tabular}
\caption{A summary of the replication models predicting z-scored pairwise cosine similarities between
embeddings of sentences with coordinated VPs from overlap scores. All coefficients are significant with 
$p < 0.001$.}\label{tab:app-coordinated-vp-overlap}
\end{table}

\subsubsection{Predicative nominals with gerund subjects}\label{app:sssec-predicative-nominals}

The following lexical items were used for the replication experiment:

\begin{itemize}
    \item Gerund subjects: \textit{proposing}, \textit{rejecting}, \textit{praising}, \textit{criticizing}
    \item Pronomial and nominal objects: \textit{him}, \textit{me}, \textit{the idea}, \textit{the design}
    \item Copula forms (same as in the original experiment): \textit{is}, \textit{was},
    \textit{will be}, \textit{is going to be}
    \item Nominal predicates: \textit{decision}, \textit{defeat}, \textit{loss}, \textit{improvement}
\end{itemize}
A summary of the replication models is shown in Tables~\ref{tab:app-gerund-subjects}.

\begin{table}[t]
\centering
\begin{tabular}{@{}lrrr@{}}
\toprule
                     & \textbf{mpnet} & \textbf{distilroberta} & \textbf{bert} \\ \midrule
\textbf{SameSubj}    & 0.82  & 0.70          & 0.31       \\
\textbf{SameCop}     & 0.35  & 0.30          & \textbf{0.55}       \\
\textbf{SameAdj}     & 0.58  & 0.79          & 0.50       \\
\textbf{SamePred}    & 0.99   & 1.01          & 0.52       \\
\textbf{SameObjNoun} & 1.01  & 1.04          & 0.60       \\
\textbf{SameObjPron} & 0.44  & 0.50          & 0.42       \\ \midrule
\textbf{R-squared}   & 0.50  & 0.54          & 0.22       \\ \bottomrule
\end{tabular}
\caption{A summary of the replication models predicting z-scored pairwise cosine similarities between
embeddings of sentences with gerund subjects and nominal predicates.
All coefficients are significant with $p < 0.001$.}\label{tab:app-gerund-subjects}
\end{table}

\subsubsection{Participant-set overlap vs.\ identical participants}

The following lexical items were used for the replication experiment:

\begin{itemize}
    \item Basic nouns: \textit{horse}, \textit{pig}, \textit{donkey}
    \item Extra nouns: \textit{elephant}, \textit{bison}, \textit{moose}
    \item Verbs: \textit{gives}, \textit{demonstrates}, \textit{entrusts}
    \item Adverbs: \textit{suddenly}, \textit{predictably}, \textit{openly}
\end{itemize}
A summary of the replication models is shown in Tables~\ref{tab:app-ditransitives}.

\begin{table}[t]
\centering
\begin{tabular}{@{}lrrr@{}}
\toprule
 & \textbf{mpnet} & \textbf{distilroberta} & \textbf{bert} \\ \midrule
\textbf{SameAdv} & 1.05 & 1.07 & 0.64 \\
\textbf{SamePred} & 0.93 & 0.64 & 0.83 \\
\textbf{Overlap} & 0.90 & 1.00 & 0.91 \\
\textbf{SPCRes} & 0.03 & 0.02 & 0.10 \\ \midrule
\textbf{R-squared} & 0.745 & 0.738 & 0.57 \\
\textbf{\begin{tabular}[c]{@{}l@{}}R-squared\\ (w/o SPCRes)\end{tabular}} & 0.744 & 0.737 & 0.56 \\ \bottomrule
\end{tabular}
\caption{A summary of the replication models predicting z-scored pairwise cosine similarities between
embeddings of sentences with ditransitive verbs. SPCRes stands for SamePosCountRes, i.e.\ the
residuals of the number of identical words in identical positions regressed on lexical overlap.
All coefficients are significant with $p < 0.001$.}\label{tab:app-ditransitives}
\end{table}

\vfill

\end{document}